\documentclass[11pt]{article}

\usepackage[preprint]{acl}

\usepackage{times}
\usepackage{latexsym}

\usepackage[T1]{fontenc}

\usepackage[utf8]{inputenc}

\usepackage{microtype}

\usepackage{inconsolata}

\usepackage{graphicx}

\usepackage{booktabs}
\usepackage{multirow} 
\title{A Stepwise-Enhanced Reasoning Framework for Large Language Models Based on External Subgraph Generation}

\author{Xin Zhang\textsuperscript{1}, Yang Cao\textsuperscript{2}, Baoxing Wu\textsuperscript{1}, Xinyi Chen\textsuperscript{2}, Kai Song\textsuperscript{2}, Siying Li\textsuperscript{1} \\
$^1$School of Information Science and Engineering, Chongqing Jiaotong University\\ $^2$School of Computer Science and Technology, Chongqing University of Posts and Telecommunications\\
}

\begin{document}
\maketitle
\begin{abstract}
Large Language Models (LLMs) have achieved strong performance across a wide range of natural language processing tasks in recent years, including machine translation, text generation, and question answering. As their applications extend to increasingly complex scenarios, however, LLMs continue to face challenges in tasks that require deep reasoning and logical inference. In particular, models trained on large scale textual corpora may incorporate noisy or irrelevant information during generation, which can lead to incorrect predictions or outputs that are inconsistent with factual knowledge.
To address this limitation, we propose a stepwise reasoning enhancement framework for LLMs based on external subgraph generation, termed SGR. The proposed framework dynamically constructs query relevant subgraphs from external knowledge bases and leverages their semantic structure to guide the reasoning process. By performing reasoning in a step by step manner over structured subgraphs, SGR reduces the influence of noisy information and improves reasoning accuracy.
Specifically, the framework first generates an external subgraph tailored to the input query, then guides the model to conduct multi step reasoning grounded in the subgraph, and finally integrates multiple reasoning paths to produce the final answer. Experimental results on multiple benchmark datasets demonstrate that SGR consistently outperforms strong baselines, indicating its effectiveness in enhancing the reasoning capabilities of LLMs.
\end{abstract}

\section{Introduction}

In recent years, the rapid development of large language models (LLMs) has marked an important milestone in the evolution of artificial intelligence. Through pretraining on massive text corpora, these models are able to generate fluent and high quality textual outputs \citep{brown2020language}. In addition, with further supervised fine tuning, they can effectively adapt to a wide range of downstream natural language processing tasks \citep{wei2022chain}. However, despite their strong performance in text generation, LLMs still face clear limitations when applied to complex reasoning scenarios. In particular, although LLMs excel at producing fluent text, they often struggle to maintain logical consistency, which can lead to outputs that deviate from real world facts. Moreover, due to their large number of parameters, the reasoning process of LLMs lacks transparency and interpretability \citep{jacovi2020towards}.

To address these challenges, an effective approach is to introduce external knowledge sources, such as knowledge graphs (KGs), into LLMs. Knowledge graphs explicitly represent structured factual knowledge, which can complement the inherent limitations of LLMs \citep{ji2022survey}. 
Previous studies have explored the use of KGs as auxiliary knowledge sources to enhance LLM reasoning capabilities \citep{lewis2020retrieval,yao2014information}. 
These methods typically inject KG derived information into the LLM by constructing prompts or providing additional contextual inputs, thereby strengthening the model reasoning process. 
Beyond knowledge graph augmentation, recent studies have explored combining large language models with symbolic or structural representations to improve systematic generalization and interpretability \citep{yang2024harnessing, yang2024can, xiong2025deliberate}.
However, most existing approaches with KG injection rely on fixed or manually designed prompt templates, and they do not fully exploit the structural information contained in knowledge graphs \citep{jiang2023structgpt,li2023tog}.

Based on this observation, we propose a stepwise enhanced reasoning framework for LLMs with external subgraph generation, abbreviated as SGR. The proposed framework leverages schema based subgraph generation from knowledge graphs to accurately extract knowledge relevant to the query. By dynamically constructing external subgraphs, SGR guides LLMs to perform step by step reasoning, thereby reducing the influence of noisy information and improving reasoning accuracy. Furthermore, the framework integrates direct reasoning and combined reasoning strategies to produce the final enhanced reasoning output.

Experimental results on multiple benchmark datasets demonstrate that the proposed method achieves superior performance compared to existing approaches. The main contributions of this work can be summarized as follows.

\begin{itemize}
    \item We propose a novel stepwise enhanced reasoning framework that improves the reasoning capability of LLMs by integrating external subgraph generation. The proposed method significantly enhances the ability of LLMs to handle complex multi step reasoning tasks.
    \item We design an interpretable knowledge enhanced reasoning process that improves the transparency and interpretability of LLM reasoning. The reasoning paths and outputs are traceable and verifiable.
    \item Extensive experiments on multiple benchmark datasets validate the effectiveness of the proposed framework and demonstrate its advantage over existing methods.
\end{itemize}

\section{Related Work}

Knowledge graphs store large amounts of external world knowledge in a structured triple format and are widely used in knowledge representation and reasoning \citep{nickel2016review}. In early studies, researchers explored methods for integrating structured knowledge into neural models \citep{sun2019bert}. Recent work has focused on enhancing LLMs by injecting KG based knowledge during training or inference, which improves performance across various tasks \citep{liu2023survey}.
Beyond static knowledge graphs, temporal knowledge graphs have been studied to capture the dynamic evolution of entities and relations over time, enabling more expressive reasoning under temporal constraints \citep{xiongtilp,xiong2024teilp, xiong2024large}.
However, directly embedding KG knowledge into LLM parameters often leads to increased training costs and reduced flexibility \citep{logan2019barack}.

To address this issue, recent studies have proposed retrieving relevant knowledge from KGs during inference and using it as additional context for LLMs \citep{lewis2020retrieval,guu2020retrieval}. These methods can be broadly categorized into retrieval enhanced methods and collaborative enhancement methods.

\subsection{Retrieval Enhanced Models}

The core idea of retrieval enhanced models is to retrieve relevant information from a knowledge source and provide it as input to the language model, thereby assisting the model in generating the final answer \citep{lewis2020retrieval}. Jiang et al. proposed a structured reasoning retrieval framework called StructGPT, which integrates iterative reading then reasoning mechanisms \citep{jiang2023structgpt}. Related approaches further explore multi hop retrieval and reasoning over structured knowledge to improve answer accuracy \citep{das2019multi,chen2020open}.

\begin{figure*}[t]
  \centering
  \includegraphics[width=0.7\linewidth]{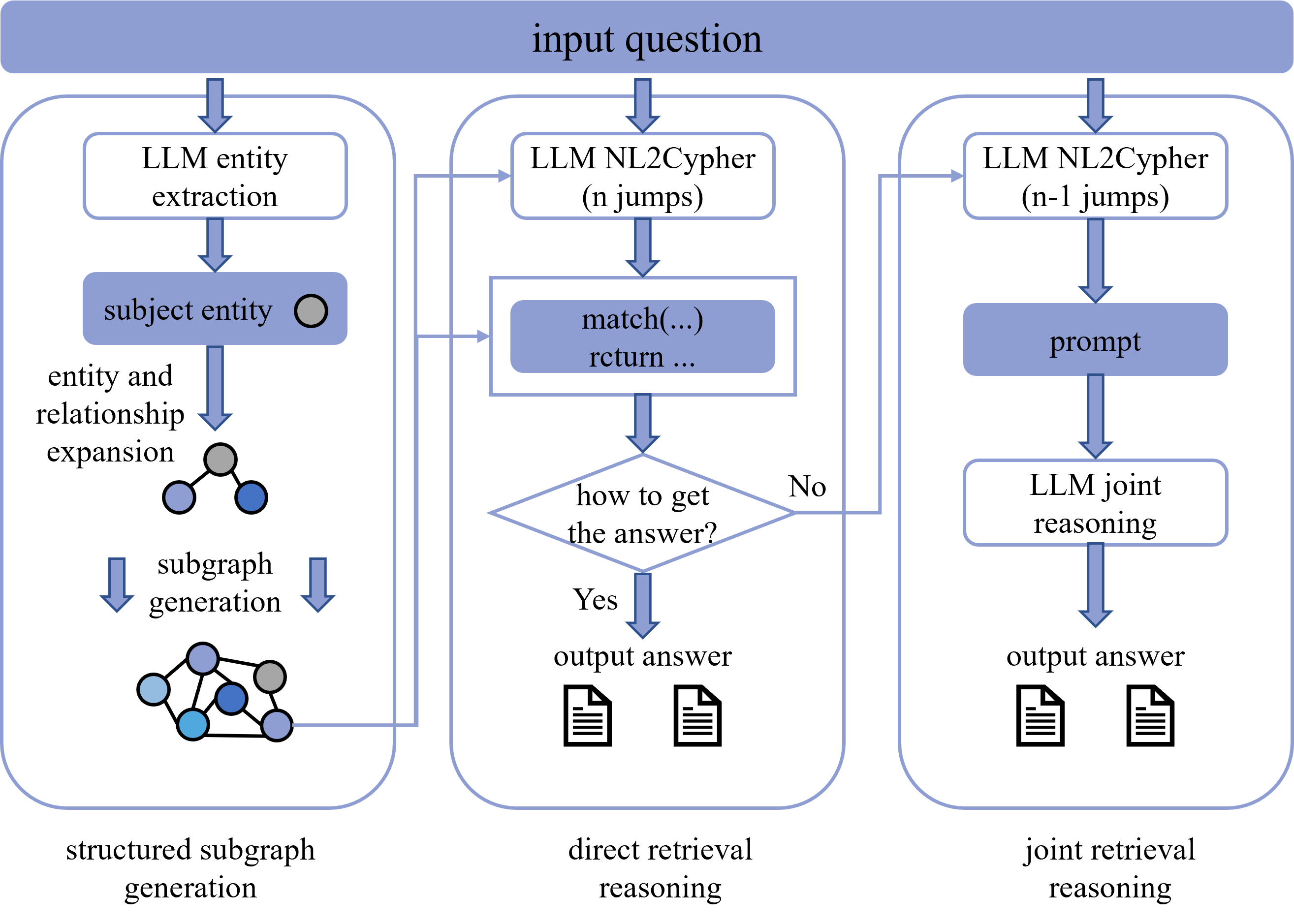}
  \caption{Pipeline of SGR framework.}
  \label{fig:fig1}
\end{figure*}

However, retrieval enhanced methods still face several limitations in practical applications. First, the retrieved information may not fully cover the structured reasoning paths provided by the knowledge graph, which can result in incomplete reasoning chains \citep{das2019multi}. Second, excessive retrieved information may introduce noise and overwhelm the model, thereby degrading reasoning performance \citep{izacard2021leveraging}. Finally, if the retrieved knowledge contains errors or redundancies, it may negatively affect the final reasoning results \citep{petroni2021kilt}.

\subsection{Collaborative Enhanced Models}

Collaborative enhanced models focus on designing an interaction mechanism between large language models and knowledge graphs. In this paradigm, the language model and the knowledge graph jointly participate in the reasoning process to generate final answers \citep{luo2022knowledge,zhou2022least}. These methods typically first generate an initial reasoning result using the language model, then leverage the knowledge graph to verify or refine the intermediate reasoning outcomes, and finally produce the enhanced answer. \citet{luo2022knowledge} proposed a KG enhanced reasoning framework in which the language model first performs reasoning on a given query, followed by knowledge validation and correction using the knowledge graph. Through iterative interactions between the language model and the knowledge graph, the system gradually improves reasoning accuracy.

In contrast to retrieval enhanced models, collaborative enhanced models emphasize bidirectional interaction between the language model and the knowledge graph. The language model not only consumes knowledge from the graph, but also generates intermediate hypotheses that guide knowledge graph queries \citep{yang2023collaborative}. Although this paradigm improves reasoning robustness, it still faces challenges related to interaction efficiency and reasoning consistency.

\section{Methodology}

To better leverage the reasoning capabilities of large language models and external knowledge graphs, we propose a stepwise enhanced reasoning framework for large language models with external subgraph generation, abbreviated as SGR. The overall framework consists of three main components, namely structured subgraph generation, direct reasoning enhancement, and collaborative reasoning enhancement.

The framework first transforms the input question into a structured query and generates a relevant external subgraph from the knowledge graph. Then, based on the generated subgraph, the language model performs stepwise reasoning to obtain candidate answers. Finally, multiple reasoning paths are integrated to produce the final output.

\subsection{Structured Subgraph Generation}

\subsubsection{Knowledge Graph Construction}

Given an input question, SGR first constructs a structured representation of the question. The language model extracts key entities, relations, and constraints from the input and maps them to corresponding elements in the knowledge graph. By identifying relevant entities and relations, the framework generates an initial subgraph that serves as the knowledge foundation for subsequent reasoning.

This structured subgraph provides explicit relational paths that guide the language model during reasoning. Compared with unstructured textual prompts, the structured subgraph enables more precise control over the reasoning process and reduces the influence of irrelevant information.

\subsubsection{Subgraph Generation Process}

The subgraph generation process consists of the following steps. First, the language model identifies the core entities and relations involved in the question. Second, the knowledge graph is queried to retrieve related entities and relations, forming candidate subgraphs. Third, irrelevant nodes and edges are filtered based on semantic relevance to the input question. The resulting subgraph contains the most relevant knowledge needed for reasoning.

By dynamically generating subgraphs tailored to each input question, SGR avoids the limitations of fixed prompt templates and fully exploits the structural information encoded in the knowledge graph.

\subsection{Stepwise Reasoning Enhancement}

Based on the generated subgraph, SGR guides the language model to perform step by step reasoning. At each step, the model focuses on a specific part of the subgraph and generates intermediate reasoning results. These intermediate results are then validated and refined using the structural constraints provided by the subgraph.

This stepwise reasoning process enables the model to decompose complex reasoning tasks into simpler subproblems. By explicitly following the relational paths in the subgraph, the model reduces logical inconsistencies and improves reasoning accuracy.

\subsection{Collaborative Reasoning Integration}

In the final stage, SGR integrates multiple reasoning paths obtained from different subgraph traversals. The language model evaluates the consistency and confidence of each reasoning path and combines them to produce the final answer. This integration process further enhances robustness and mitigates the impact of potential errors in individual reasoning paths.

Overall, the proposed framework provides a unified approach that combines structured knowledge, stepwise reasoning, and collaborative integration to improve the reasoning capabilities of large language models.

The model first performs entity based reasoning to generate a structured subgraph schema. The schema serves as a representation of external knowledge relevant to the input query.

The language model is instructed to act as a knowledge graph expert and is provided with several knowledge graph triples. It is required to reason over these triples and generate a structured query schema. Specifically, the model extracts core entities and relations from the input question, and then organizes them into a schema that reflects the logical structure of the reasoning process. An example is shown as follows. Given the question, the model identifies entities and relations such as entity one relation entity two, entity two relation entity three, and entity three relation entity four. The generated schema consists of a set of entity relation pairs that capture the reasoning path required to answer the question.

During this process, the SGR framework leverages schema based reasoning to constrain the search space of knowledge graph queries. The generated schema is then used to guide the construction of a structured query in Neo4j, enabling accurate retrieval of relevant knowledge.

\begin{table*}[t]
\centering
\small
\caption{Performance comparison of different reasoning methods on CWQ, WebQSP, and GrailQA. 
Hits@1 and accuracy are reported where applicable, and the best results for each metric are highlighted in bold.
Note: Best results are taken from prior work, including $\alpha$,
$\beta$,
$\gamma$,
and $\delta$.}
\label{tab:experimental_results}
\begin{tabular}{lcccccc}
\toprule
\multirow{2}{*}{Method} &
\multicolumn{2}{c}{CWQ} &
\multicolumn{2}{c}{WebQSP} &
\multicolumn{2}{c}{GrailQA} \\
\cmidrule(lr){2-3} \cmidrule(lr){4-5} \cmidrule(lr){6-7}
& Hits@1 & Acc & Hits@1 & Acc & Hits@1 & Acc \\
\midrule
IO Prompt / ChatGPT        & 0.376 & 0.256 & 0.633 & 0.582 & 0.294 & 0.223 \\
CoT / ChatGPT              & 0.388 & 0.258 & 0.622 & 0.577 & 0.281 & 0.201 \\
\midrule
Prior FT SOTA              & 0.704$^{\alpha}$ & -- & 0.821$^{\beta}$ & -- & 0.754$^{\gamma}$ & -- \\
Prior Prompting SOTA       & -- & -- & 0.744$^{\delta}$ & -- & 0.532$^{\delta}$ & -- \\
\midrule
SGR / Cypher LLM         & 0.523 & 0.445 & 0.745 & 0.706 & 0.624 & 0.593 \\
SGR / ChatGPT            & 0.578 & 0.526 & 0.801 & 0.784 & 0.713 & 0.633 \\
StructGPT / ChatGPT        & -- & -- & 0.726 & -- & -- & -- \\
ToG / ChatGPT              & 0.571 & -- & 0.762 & -- & 0.687 & -- \\
ToG / GPT-4                & \textbf{0.725} & -- & 0.826 & -- & \textbf{0.814} & -- \\
SGR / GPT-4              & 0.632 & \textbf{0.590} & \textbf{0.826} & \textbf{0.808} & 0.756 & \textbf{0.703} \\
\bottomrule
\end{tabular}

\end{table*}

\subsection{Direct Reasoning Enhancement}

\subsubsection{Cypher LLM}

In the direct reasoning enhancement stage, the framework first constructs a Cypher query based on the generated schema. To ensure correctness and efficiency, we employ NLCypher as the language interface for generating Cypher queries. NLCypher maps natural language questions into Cypher queries, which can be executed directly on the Neo4j database.

The Cypher queries generated by NLCypher are executed on the knowledge graph to retrieve candidate answers. In our experiments, we use a total of 140 thousand knowledge graph queries across multiple datasets. The retrieved results are then combined with the original question and provided to the large language model as structured evidence to support answer generation.

We evaluate the effectiveness of Cypher based reasoning using GPT based models. Experimental results show that the Cypher driven reasoning process significantly improves the accuracy of question answering, especially in complex multi hop reasoning scenarios.

\subsubsection{Answer Validation}

After obtaining candidate answers from the Cypher queries, the framework performs answer validation by comparing the generated answers with the retrieved knowledge. If the candidate answers match the expected schema constraints, they are accepted as valid results. Otherwise, the framework triggers further reasoning steps to refine the answers.

This validation process reduces the risk of hallucinated outputs and ensures consistency between the language model predictions and the underlying knowledge graph.

\subsection{Collaborative Reasoning Enhancement}

In cases where direct reasoning fails to produce correct answers, the framework employs collaborative reasoning enhancement. In this stage, the language model and the knowledge graph interact iteratively to refine the reasoning process. The language model generates intermediate hypotheses, which are then verified and expanded using the knowledge graph. The verified results are fed back into the language model for further reasoning.

Through iterative collaboration, the framework explores multiple reasoning paths and gradually converges to a more accurate answer. This approach improves robustness in scenarios where a single reasoning pass is insufficient.

\section{Experiments}

\subsection{Experimental Setup}

We evaluate the proposed framework on several benchmark datasets, including CWQ, WebQSP, GrailQA, and KQA Pro. CWQ is a complex question answering dataset that focuses on multi hop reasoning over knowledge graphs. WebQSP is a widely used benchmark for question answering over Freebase. GrailQA emphasizes generalization to unseen query structures. KQA Pro is a knowledge based question answering dataset with diverse reasoning types.

For evaluation, we use standard metrics including accuracy and F1 score. Accuracy measures the proportion of questions for which the model produces the correct answer, while F1 score captures the balance between precision and recall.

\begin{figure}[t]
  \centering
  \includegraphics[width=0.9\linewidth]{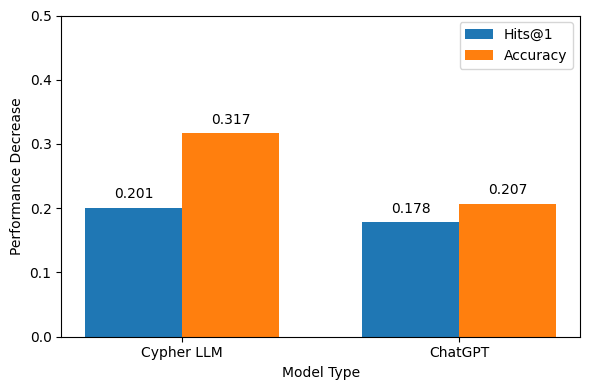}
  \vspace{-10pt}
  \caption{Impact brought by removing Schema prompts.}
  \includegraphics[width=0.9\linewidth]{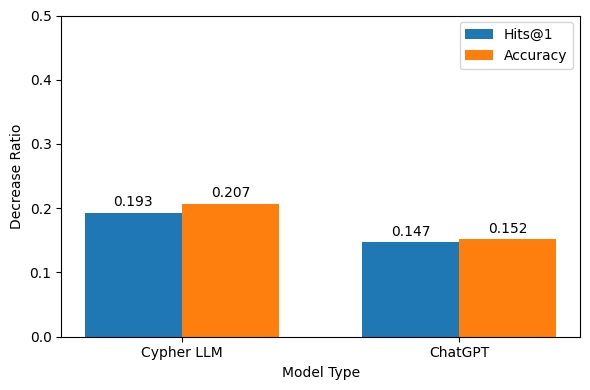}
  \vspace{-10pt}
  \caption{Impact brought by removing neo4j retrieval.}
\end{figure}

\begin{table*}[t]
\centering
\small
\caption{Ablation experiment results on the CWQ dataset}
\label{tab:ablation_cwq}
\begin{tabular}{lcccccccc}
\toprule
\multirow{2}{*}{Method} &
\multicolumn{2}{c}{With Schema} &
\multicolumn{2}{c}{With neo4j} &
\multicolumn{2}{c}{Without Schema} &
\multicolumn{2}{c}{Without neo4j} \\
\cmidrule(lr){2-3} \cmidrule(lr){4-5} \cmidrule(lr){6-7} \cmidrule(lr){8-9}
 & Hits@1 & Acc & Hits@1 & Acc & Hits@1 & Acc & Hits@1 & Acc \\
\midrule
SGR/Cypher LLM & 0.523 & 0.445 & 0.553 & 0.445 & 0.322 & 0.128 & 0.360 & 0.238 \\
SGR/ChatGPT   & 0.578 & 0.526 & 0.578 & 0.526 & 0.400 & 0.319 & 0.431 & 0.374 \\
\bottomrule
\end{tabular}
\end{table*}

\subsection{Experimental Results}

To evaluate the performance of the proposed method, this paper conducts extensive experiments on two representative datasets, namely CWQ and GrailQA, using input output style prompting and chain of thought reasoning. The experiments compare SGR with several strong baselines, including ChatGPT, CoT based prompting, and prior state of the art methods.

First, to investigate the impact of external knowledge, we compare the performance of SGR under different settings. Experimental results show that incorporating external subgraph knowledge significantly improves the reasoning performance of large language models. Compared with methods that rely solely on textual prompts, SGR achieves higher accuracy and Hits@1 scores across all datasets, demonstrating the effectiveness of external subgraph generation.

In addition, we compare SGR with prior state of the art reasoning methods. The results indicate that SGR consistently outperforms existing approaches on both CWQ and GrailQA. Notably, SGR achieves substantial improvements even when applied to smaller language models, suggesting that the proposed framework effectively compensates for the limited reasoning capacity of smaller models by leveraging structured external knowledge.

Furthermore, we analyze the generalization ability of SGR across different base models. Experimental results show that SGR yields consistent performance gains when applied to both Cypher LLM and ChatGPT, indicating that the framework is model agnostic and can be easily integrated with various large language models.

\subsection{Ablation Study}

To further analyze the contribution of each component in the proposed framework, we conduct ablation studies on the CWQ dataset. Specifically, we evaluate the impact of removing schema based subgraph generation and neo4j based retrieval, respectively.

The results show that removing schema based prompts leads to a noticeable performance degradation, confirming the importance of schema guidance in constructing accurate subgraphs. Similarly, eliminating neo4j based retrieval significantly reduces reasoning accuracy, indicating that structured knowledge retrieval plays a critical role in supporting multi step reasoning.

These findings demonstrate that both schema guidance and structured retrieval are essential components of SGR, and their combination is crucial for achieving optimal reasoning performance.

\subsection{Application Scenarios and Error Analysis}

SGR can be applied to a wide range of knowledge intensive reasoning tasks, including complex question answering and logical inference. In practical applications, SGR exhibits strong robustness and stability, particularly in scenarios that require multi step reasoning over structured knowledge.

We further conduct an error analysis to identify the main sources of remaining errors. The analysis reveals that most errors arise from incomplete subgraph construction and ambiguous entity linking. In some cases, missing or noisy knowledge in the external knowledge graph also leads to incorrect reasoning paths. Addressing these issues by improving subgraph generation quality and knowledge coverage is a promising direction for future work.

To further extend the applicability of the proposed framework, we explore its use in the medical domain by integrating medical knowledge graphs with LLMs to support complex reasoning tasks such as disease diagnosis and clinical decision making. By constructing and leveraging structured medical knowledge graphs, the SGR framework assists LLMs in reasoning over medical facts, symptoms, and treatment relations, thereby improving reasoning accuracy and reliability in real world medical scenarios.

To validate the effectiveness of SGR in different domains, we conduct experiments on academic knowledge graph based reasoning tasks. The experimental results demonstrate that SGR consistently outperforms baseline methods across multiple datasets. In addition, by dynamically generating external subgraphs, the framework enables LLMs to flexibly adapt their reasoning strategies under different contexts, thereby achieving more precise reasoning and inference.

\section{Conclusion}

This paper proposes a stepwise enhanced reasoning framework for large language models based on external subgraph generation. By dynamically constructing relevant subgraphs from knowledge graphs, the proposed method accurately extracts structured knowledge and guides LLMs to perform step by step reasoning. Through the integration of direct reasoning and combined reasoning strategies, SGR effectively improves the accuracy and reliability of reasoning results. Extensive experimental results across multiple benchmark datasets demonstrate the strong reasoning enhancement capability and generalization ability of the proposed framework. Moreover, SGR improves the interpretability of the reasoning process by providing explicit reasoning paths supported by external knowledge.

\section*{Limitations}

Despite its effectiveness, the proposed framework has several limitations. In particular, SGR introduces additional computational overhead due to subgraph construction and retrieval, which may affect inference efficiency in large scale applications. In addition, the performance of the framework depends on the quality and coverage of the underlying knowledge graphs, and incomplete or noisy knowledge may limit its effectiveness. Future work will focus on optimizing the efficiency of external subgraph generation and retrieval, as well as exploring lightweight deployment strategies to further enhance the practicality of the framework in real world scenarios.

\bibliography{custom}

\end{document}